\documentclass[10pt,twocolumn,letterpaper]{article}

\usepackage[pagenumbers]{cvpr} 

\usepackage[dvipsnames]{xcolor}
\usepackage{algorithm2e}

\definecolor{cvprblue}{rgb}{0.21,0.49,0.74}
\usepackage[pagebackref,breaklinks,colorlinks,citecolor=cvprblue]{hyperref}
\usepackage[normalem]{ulem} 

\def\paperID{16760} 
\def\confName{CVPR}
\def\confYear{2024}

\title{Decentralised Traffic Incident Detection via Network Lasso}

\author{
Qiyuan Zhu\\
Swinburne University of Technology\\
Melbourne, Australia
\and
A. K. Qin\\
Swinburne University of Technology\\
Melbourne, Australia
\and
Prabath Abeysekara\\
Swinburne University of Technology\\
Melbourne, Australia
\and
Hussein Dia\\
Swinburne University of Technology\\
Melbourne, Australia
\and
Hanna Grzybowska\\
CSIRO, Sydney, Australia
}

\begin{document}
\maketitle
\begin{abstract}
Traffic incident detection plays a key role in intelligent transportation systems, which has gained great attention in transport engineering. In the past, traditional machine learning (ML) based detection methods achieved good performance under a centralised computing paradigm, where all data are transmitted to a central server for building ML models therein. Nowadays, deep neural networks based federated learning (FL) has become a mainstream detection approach to enable the model training in a decentralised manner while warranting local data governance. Such neural networks-centred techniques, however, have overshadowed the utility of well-established ML-based detection methods. In this work, we aim to explore the potential of potent conventional ML-based detection models in modern traffic scenarios featured by distributed data. We leverage an elegant but less explored distributed optimisation framework named Network Lasso, with guaranteed global convergence for convex problem formulations, integrate the potent convex ML model with it, and compare it with centralised learning, local learning, and federated learning methods atop a well-known traffic incident detection dataset. Experimental results show that the proposed network lasso-based approach provides a promising alternative to the FL-based approach in data-decentralised traffic scenarios, with a strong convergence guarantee while rekindling the significance of conventional ML-based detection methods.
\end{abstract}    
\section{Introduction}
\label{sec:intro}

Automatic incident detection (AID) plays a crucial role in intelligent transportation systems (ITS), aiming to enhance the efficiency of traffic management \cite{IOT}. It is capable to detect traffic incidents (e.g., car accidents and unexpected disruptions) via traffic data analysis and modelling, and thus allow for making prompt and effective responses to mitigate the incidents occurred. 

AID is often challenged by the complex inherent characteristics of traffic data. For example, there may exist some traffic patterns which resemble those featured by incidents, e.g., in the presence of compression waves, roadway bottlenecks, and equipment malfunctions. Discontinuities in the traffic flow, introduced by vehicles entering and exiting from side streets, can also produce incident-like traffic patterns. 
Due to the existence of such patterns, AID may lead to an undesired amount of false alarms. \cite{review1}.
Moreover, traffic incidents are sparsely occurring events and may demonstrate a large variety of forms. Therefore, it is typical that the availability of non-incident data overwhelms that of incident data. Even if their amounts are equivalent, the incident data are often under-represented due to the huge variations of incidents. Such imbalanced and unrepresentative issues pose a great challenge to detection techniques \cite{ANN1} and have made significant advancements through machine learning (ML) techniques \cite{Cal1, SND1, SND2, AIDA1, RF1, SVM1}.

On the other hand, the continual expansion of transportation networks and the subsequent upsurge in traffic volumes have brought forth new challenges that current approaches struggle to effectively address. 
Most existing practical applications typically train these AID modes using centralised computing paradigm (e.g. in cloud-based computing infrastructure). \textit{This requires all traffic data to be transmitted to a central server, necessitating substantial bandwidth for data transmission}.
Although ML-based techniques are exceptionally lightweight and computationally efficient, the continual expansion of the transportation network and the increasing volume of traffic data have led to escalating computational and storage costs, as well as inevitable detection delays \cite{review1}. 
Furthermore, centralised \textit{one-size-fits-all} models trained using machine learning and neural network techniques on data from diverse traffic regions with different traffic characteristics often exhibit subpar generalization performance \cite{review1, review3, RN1}. Consequently, the detection model trained in one location may not be generaliseable to other locations, leading to a diminished capabilities of AID method \cite{review4}.


Nowadays, federated learning (FL) \cite{Fed1, Fed2} has emerged as the prevailing approach for traffic incident detection. This approach warrants local data governance and enables swift and precise detection in a data-decentralised manner. However, \textit{FL is primarily tailored for deep learning methods, which has overshadowed the utility of well-established machine learning-based detection techniques}. Besides, FL can diverge with data residing on different location and devices \cite{Fed6}. There exists another lesser-known yet equally potent distributed optimisation framework Network Lasso (NL), which seamlessly fits the aforementioned problem setting \cite{NL1, NL2}. NL has the capability to concurrently cluster and optimise a family of context-aware detection models with guaranteed global convergence, which is one of the demanded characteristics for traffic management. It also allows data processing and analysis to occur in proximity to where traffic data is collected (i.e. edge servers), subsequently reducing the load on the existing infrastructure \cite{NL2}. In addition, NL also enables us to \textit{reuse suitable existing ML approaches with minimal adjustments}, leading to a compelling synergy between machine learning and traffic expertise. This facilitates the \textit{swift adoption of established ML-based AID strategies into decentralised traffic networks}. 


Motivated by the merits of the NL framework to tackle the aforementioned challenges, we aim to explore its potential in detecting traffic incidents within decentralised modern transportation networks - \textit{a problem yet unexplored}. Our proposed approach harnesses NL to parallelly training a distributed family of one-class SVMs \cite{OCSVM1}, which is a prevalent approach for traffic incident detection. The key contributions of our paper are summarised below:
\begin{enumerate}
\item{We leverage an elegant but less explored distributed optimisation framework named Network Lasso, with guaranteed global convergence for convex problem formulations, to rekindle the potential of conventional ML-based AID methods in modern traffic scenarios featured by data decentralisation.}
\item{We construct the graph in NL by considering both the geo-spatial proximity and road network geometry of the traffic regions (nodes) so that the knowledge sharing is enabled across similar traffic regions in terms of distance and road geometry which are connected in the graph.} 
\item{We comprehensively evaluated the proposed approach to tackle the challenges outlined, and compare it against the representative centralised, localised, and federated learning methods. The results show that the proposed approach outperforms centralised and localised learning and provides a promising alternative to federated learning in data-decentralised traffic scenarios.}
\end{enumerate}

The remainder of this paper is organised as follows. \cref{sec:related-work} provides a brief summary of the related traffic AID works, before discussing the necessary preliminaries and proposed solution in \cref{sec:Preliminaries} and \cref{sec:Methods}, respectively. \cref{sec:experiment} introduces the experimental settings and discusses the results. \cref{sec:conclusion} concludes the paper with future work mentioned.

\section{Related work}
\label{sec:related-work}

This section briefly reviews the existing ML-based centralised as well as decentralised traffic AID approaches.

\subsection{Traffic incident detection}
Detecting roadway incidents typically involves analysing a time-series of traffic records and assigning a label to each record to indicate the presence or absence of an incident. When incident labels are missing or lacking of quality \cite{OCC2}, an alternative approach is to measure an anomaly score against each data instance, followed by making decisions based on a threshold to distinguish the incident event from the rest. \textit{In this study, our emphasis is solely on score-based detection, utilising single-variate time-series as the input}.

Due to the availability of traffic surveillance systems, the most intuitive approach to detect incidents involves analysing general traffic information to differentiate abnormal events from routine traffic occurrences. The well-known California algorithm \cite{Cal1, Cal2} and related approaches employ pattern analysis, detecting incidents by monitoring fluctuations in occupancy measurements along road sections. 
However, the California algorithm operates case-wisely, limiting its broader applicability. To address this, the Standard Normal Deviate (SND) Algorithm \cite{SND1, SND2} computes mean values from historical traffic data, triggering alarms upon significant deviations. While these methods achieve higher detection rates, \textit{their sensitivity to outliers lead to inconsistent performance.}

\subsection{Machine learning-based centralised AID}

Machine learning techniques have been used in the existing literature to learn the data representation and differentiate incidents. For example, Yang \etal in \cite{AE1} proposed an autoencoder-based incident detection framework. Their method utilises a spatio-temporal dataset containing several months of incident-free data, leveraging the necessity of labelled incidents. In \cite{OCSVM1}, Amraee \etal employed a one-class SVM (OC-SVM) and the histogram of optical flow (HOF) to detect abnormal traffic events. Another traffic anomaly detection model in \cite{OCSVM2} attempted the detection of traffic anomalies using the Informer and OC-SVM techniques. \textit{These unsupervised approaches based on OC-SVM have demonstrated their ability to strike a balance between false alarms and missed alarms, even in the absence of incident data}.

Meanwhile, Deep Learning (DL) techniques have also commonly been used in most recent approaches \cite{CNN2, LSTM1}. However, the challenge lies in real traffic data being predominantly non-incidents, making it difficult for the networks to distinguish and classify incidents accurately \cite{OCC2, GAN2_SVM}. Research has focused on sampling techniques \cite{PNN1} and deep generative techniques \cite{GAN2_SVM} with some success. Despite those progress, \textit{the computational demands remain burdensome for resource-constrained decentralised deployments in current transportation infrastructures}.

\subsection{Decentralised learning-based AID}
Recently, the concept of smart cities has ignited a surge in research on distributed traffic scenarios \cite{Fed1, Fed2, Fed3, Fed4, Fed5}. The majority of research conducted on data-decentralised incident detection scenario leans towards DL-based approaches, owing to the inherent compatibility between Federated Learning (FL) techniques and parameterised DL methods \cite{Fed2, Fed5}. \textit{Considering the scalability challenges faced by existing centralised AID methods} in the face of expanding traffic networks, and \textit{distributed DL methods often incurring substantial costs}, \textit{there is a strong motivation towards extending the capabilities of centralised ML-based AID methods into distributed scenarios}. To that end, we explore the ability of NL to train a family of OC-SVMs co-optimised using ADMM \cite{ADMM}, employing incident-free samples to leverage the necessity of incident data.
\section{Preliminaries}
\label{sec:Preliminaries}

This section provides a brief overview of the technical preliminaries associated with the proposed work.

\subsection{One-class support vector machine}
One-class classification (OCC) \cite{OCC1} represents a unique type of multi- or binary-classification task in which the focus is solely on instances belonging to a single class. The traffic AID problem can be effectively characterised as an OCC problem and solved by an one-class classifier, such as one-class SVM (OC-SVM) \cite{OCSVM1}. Given a set of N incident-free samples, denoted as ${X}_{Train} = {x_1, x_2,...,x_N}$, the objective for OC-SVM is to find a boundary, which determined by its kernel function to encapsulate training samples. Anything separated by the boundary is considered as an incident which doesn't follow the normal traffic's characteristics.  
 
To train the OC-SVM, we ascertain the weight parameter $w$ by minimising the optimisation objective \cite{OCC2, OCC3}:

\begin{equation}
\label{eq:1}
\begin{aligned}
    \min_{w,\xi,b} \quad & \frac{1}{2} {\lVert w \rVert}^{2} + \frac{1}{\nu N} \sum_{i=1}^{N}{\xi_{i}} - {b}\\
    \textrm{s.t.} \quad & \langle w,\phi(x_{i}) \rangle \geq {b} -\xi_{i}, \xi_{i}\geq0    \\
\end{aligned}
\end{equation}

where, the column vector $\xi = [\xi_{1} , \xi_{2}, . . . , \xi_{N} ]$ with each $\xi_{i}$ representing the slack variable corresponding to the \textit{i}-th training sample (i.e., time series of occupancy difference). The function $\phi$ maps $x_i$ to a kernel space where dot products are defined through a kernel function $K(\cdot , \cdot)$. Additionally, we have the bias term denoted as '$b$', a trade-off parameter '$\nu $', and '$N$' represents the number of training examples. Once the optimisation is solved, detecting incidents in a query sample ${x}_{test}$ can be accomplished using the decision function on anomaly score by:

\begin{equation}
\label{eq:2}
\begin{aligned}
    \textrm{sgn}(\langle w,\phi(x_{test}) \rangle - {b})
\end{aligned}
\end{equation}

\subsection{Network lasso}

\begin{figure*}[t!]
  \centering
  \begin{subfigure}[t]{0.32\textwidth}
    \includegraphics[width=\textwidth]{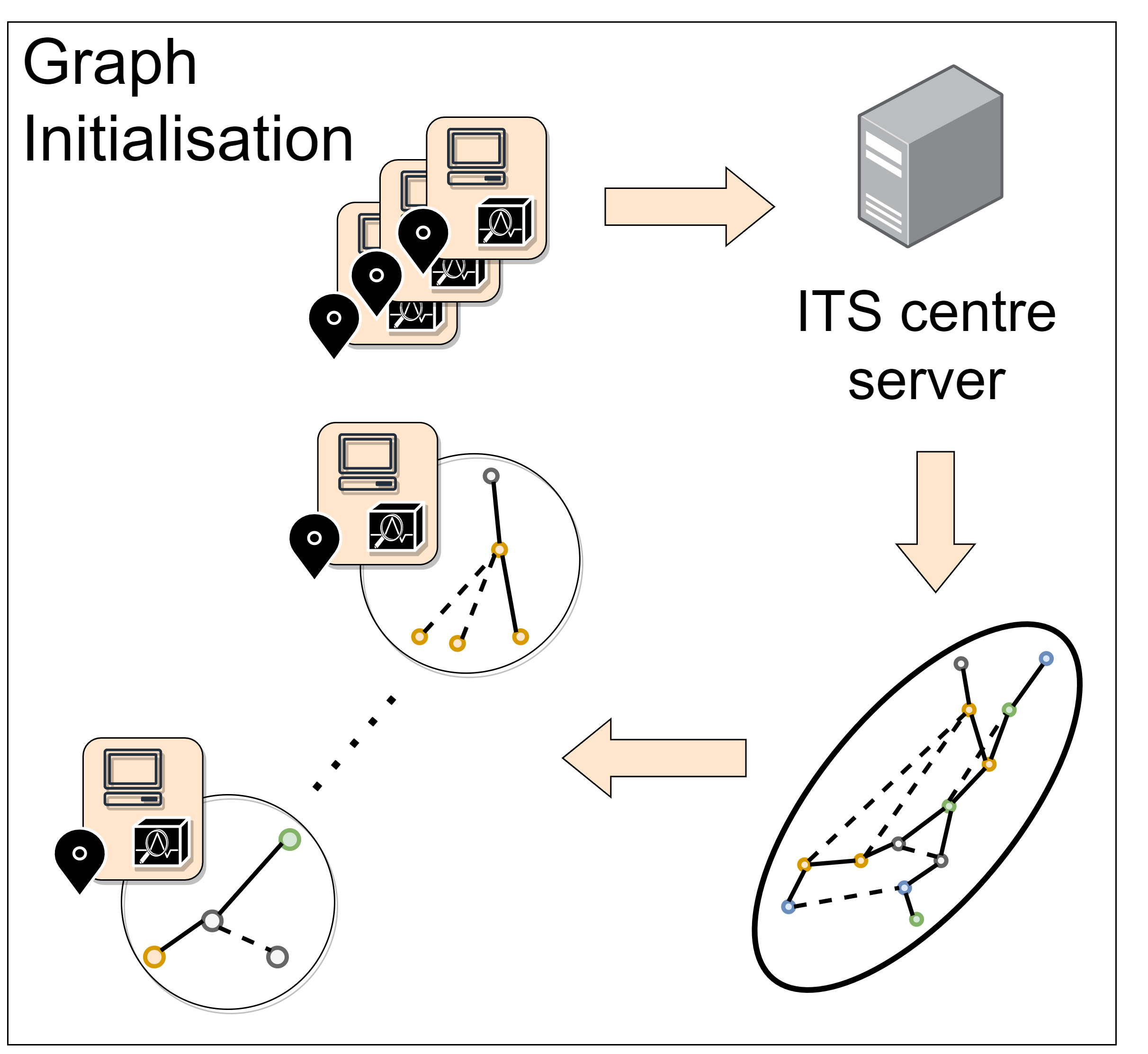}
    \caption{The initialisation of NL graph.}
    \label{fig:framework-1}
  \end{subfigure}
  \hfill
  \begin{subfigure}[t]{0.32\textwidth}
    \includegraphics[width=\textwidth]{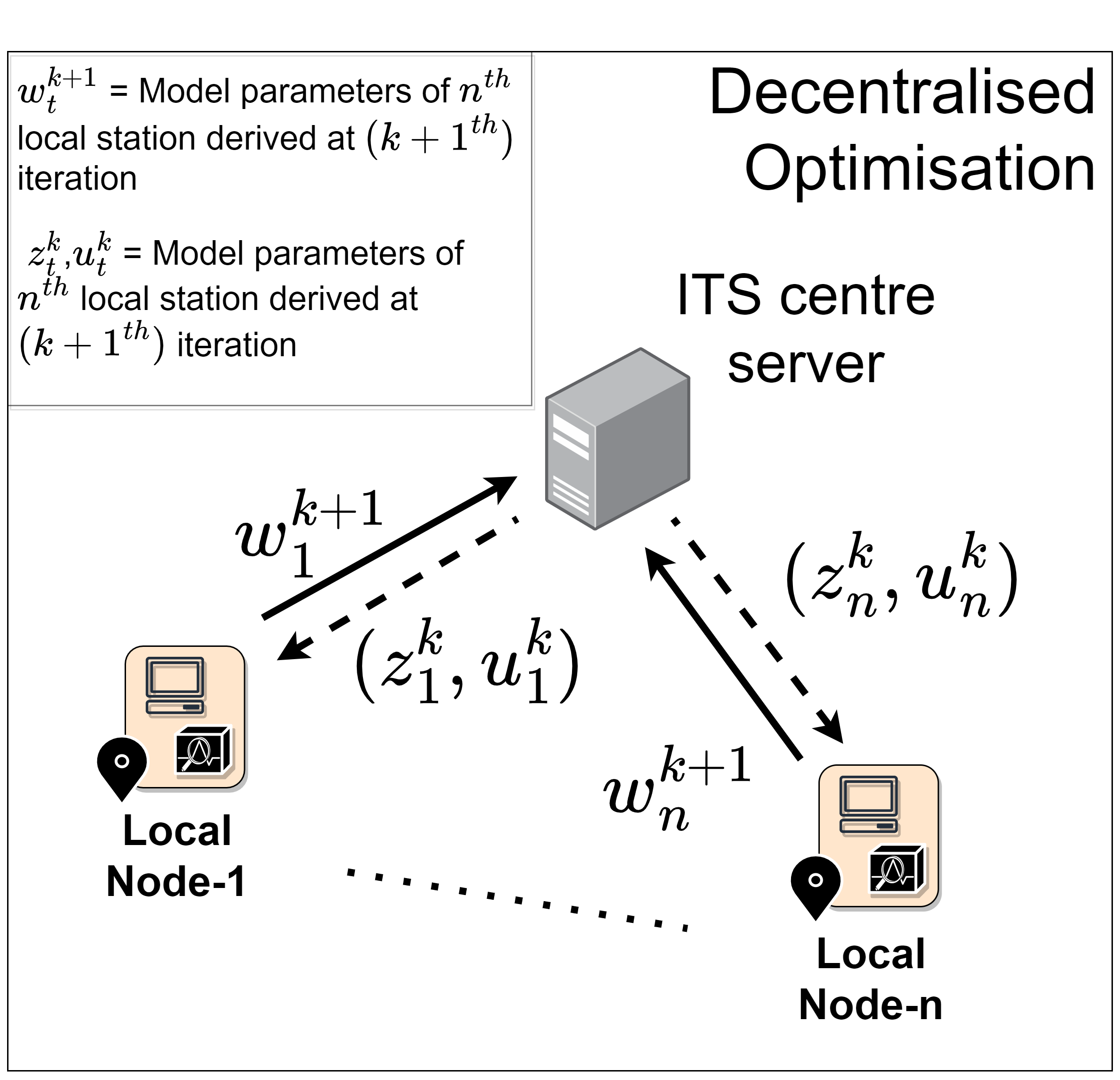}
    \caption{The information flow in NL over a hierarchical client-server framework.}
    \label{fig:framework-2}
  \end{subfigure}
  \hfill
  \begin{subfigure}[t]{0.32\textwidth}
    \includegraphics[width=\textwidth]{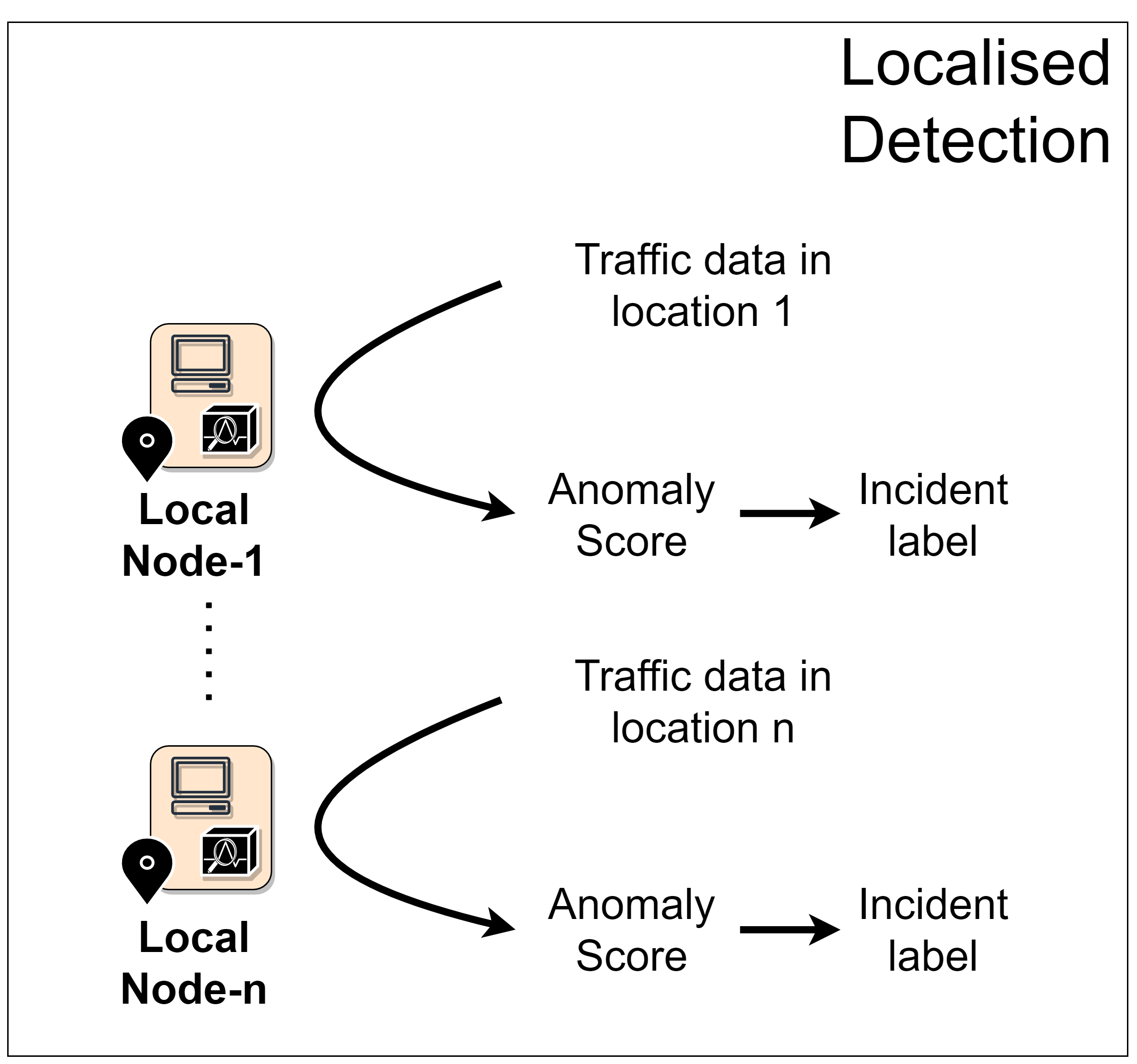}
    \caption{The detection of incidents that happened in local nodes.}
    \label{fig:framework-3}
  \end{subfigure}
  \caption{The NL-based framework for detecting traffic incidents on modern traffic scenarios featured by data-decentralisation.}
  \label{fig:framework-whole}
\end{figure*}

\label{sec:Preliminaries-NL}
Network Lasso (NL) is a framework designed to address large-scale optimisation problems formulated within a graph structure, enabling concurrent decentralised clustering and optimisation \cite{NL1}. In the context of an undirected networked graph, denoted as $G = (V, E)$, where nodes are represented as $V = {1, . . . , N}$, and their interconnections are established via edges, denoted as $E = \{(v_1, v_2) : v_1, v_2 \in V, v_1 \neq v_2\}$, the mathematical expression of the NL problem is outlined as follows.

\begin{equation}
\label{eq:3}
    \textrm{minimise} \quad \sum_{t \in V}{l_t 
 (w_t)} + \lambda \sum_{(j,k) \in E}{{a}_{jk} {\lVert w_j - w_k \rVert}_{2}}\\   
\end{equation}

In this optimisation problem, $w_t \in {\mathbb{R}}^{n}$ represents the model parameters of a convex loss function $l_t$. Each loss function, denoted as $l_t$, is defined over the input-output space $l_t: {\mathbb{R}}^{n} \rightarrow {\mathbb{R}} \cup \{\infty\} $ and is localised to a node $v_t \in V $ within the graph $G$. These individual loss functions serve the purpose of estimating the model parameters for each node in the graph, $v_t \in V$, through the formulation of the optimisation problem.

The parameter $\lambda$ plays a pivotal role as a regularisation factor, serving to adjust the significance of the edge objectives against the node objectives. The symbol ${a}_{jk}$ signifies the relational coefficients of a specific edge, such as $(v_j, v_k)$, on the finite-sum problem computed over the loss functions of all participating nodes in the optimisation problem. $w_j$ and $w_k$ correspond to the parameters of the models associated with two adjacent nodes $v_j$ and $v_k$ in the graph, respectively. In addition, the regularisation factor $\lambda$, the relational coefficients ${a}_{jk}$, and the L2-norm computed over the difference of model parameters between the two connected nodes $(v_j, v_k)$ collectively form a penalty factor. This penalty factor enforces the similarity of model parameters between the two connected nodes, effectively enhancing the cohesion among nodes that share similar model parameters, such that $w_j = w_k$. Consequently, nodes with congruent data distributions gravitate towards forming clusters, while those with divergent data distributions evolve into distinct clusters.

\RestyleAlgo{ruled}
\SetKwComment{Comment}{/* }{ */}
\SetKwInput{kwReq}{Require}
\begin{algorithm}
\caption{NL parallelised by ADMM}\label{alg:1}
\kwReq{${\epsilon}^{primal}$, $\ {\epsilon}^{dual}$}
\While{${\lVert {r}_{p}^{k} \rVert}_{2} < {\epsilon}^{primal} \  and \ {\lVert {r}_{s}^{k} \rVert}_{2} < {\epsilon}^{dual}$}{
${w}_{t}^{k+1} = \underset{{w}_{t}}{\arg\min}\{{l_t(w_t)}+\underset{i \in N(t)}{\sum}{\frac{\rho}{2}{\lVert {w}_{t} - {z}_{ti}^{k} - {u}_{ti}^{k}  \rVert}_{2}^{2}} \}$
${z}_{ti}^{k+1} = \theta({w}_{t} + {u}_{ti}) +(1-\theta)({w}_{i} + {u}_{ti})$
${z}_{it}^{k+1} = (1-\theta)({w}_{t} + {u}_{ti}) + \theta({w}_{i} + {u}_{ti}) $
${u}_{ti}^{k+1} = {u}_{ti}^{k} + ({w}_{t}^{k+1} - {z}_{ti}^{k+1})$ 
}
\end{algorithm}

The ADMM method is the default approach for decomposing problem \cref{eq:3} into smaller sub-problems. Within this method, each task autonomously addresses its own data-local sub-problem in parallel, shares its solution with neighboring nodes, and iterates through this process until the entire network converge. This procedural framework leads to the parallel algorithm depicted in \cref{alg:1} in which ${\epsilon}^{primal}$ and ${\epsilon}^{dual}$ correspond to the primal and dual residuals, which are typically used to enforce the stopping criteria \cite{ADMM}.

\section{Proposed Solution}
\label{sec:Methods}

This section provides a comprehensive overview of the proposed solution.

\subsection{Traffic AID as a decentralised OCC problem}

In our work, we consider a data-decentralised traffic scenario based on \cite{Cal1, review1}, see Figure \ref{fig:pems}. We consider that the selected area consists of multiple \textit{traffic regions} with different traffic characteristics. These distinctive traffic characteristics form different \textit{context environments} within the traffic regions may overlap based on the similarity. We further assume that there exists \textit{an edge-server in close proximity to a traffic region} enveloped between a upstream section and a downstream section of the road network that accumulates traffic information collected through suitable loop detectors. We call these edge-servers \textit{local nodes}, which would also provide computing resources for model training and incident detection. Meanwhile, a \textit{centre server} that runs within a centralised cloud-based infrastructure of an ITS facilitates managing the information shared during parallel optimisation, the details of which will be discussed in \cref{subsec:decentralised-traffic-detection}. For simplicity, the communication delay or failures that may occur amongst nodes participating in distributed model training as well as road sections were not considered.

To identify incidents occurring within the \textit{coverage area of a local node}, we extract time series data from the traffic region, and then approach the incident detection problem as an OCC problem, primarily for two compelling reasons. Firstly, the incident samples are either unavailable or insufficient in number to train a conventional, non-OCC classifier. In numerous cases, the collected samples fall short in meeting the required quantity \cite{OCC1}. Secondly, OCC models demand considerably fewer computational resources compared to deep learning models, making them suitable for deployment on resource-constrained distributed devices.

As depicted in \cref{fig:framework-whole}, the end-to-end solution we propose can be delineated into three stages: (1) graph initialisation, (2) decentralised optimisation, and (3) localised incident detection. In first stage, we integrate ideas from prior research on traffic POIs \cite{RN1}, and \textit{propose a strategy that combines geometric characteristics with road network into the graph design}. Because Overly-simplified road network-based graph designs often yield unsatisfactory results due to their inability to properly represent real-world traffic networks and their characteristics. In second stage, we incorporate the fused graph into NL using OC-SVM as the \textit{local classifier, which is trained atop the locally accumulated traffic information within a local node}. In third stage, the detection of incidents will be completed in local nodes. In the subsequent sections, we will introduce the traffic graph construction strategy and elaborate the NL-based decentralised OC-SVM optimisation framework that been proposed in our work.

\begin{figure}[h!]
  \centering
  \includegraphics[width=1.0\linewidth]{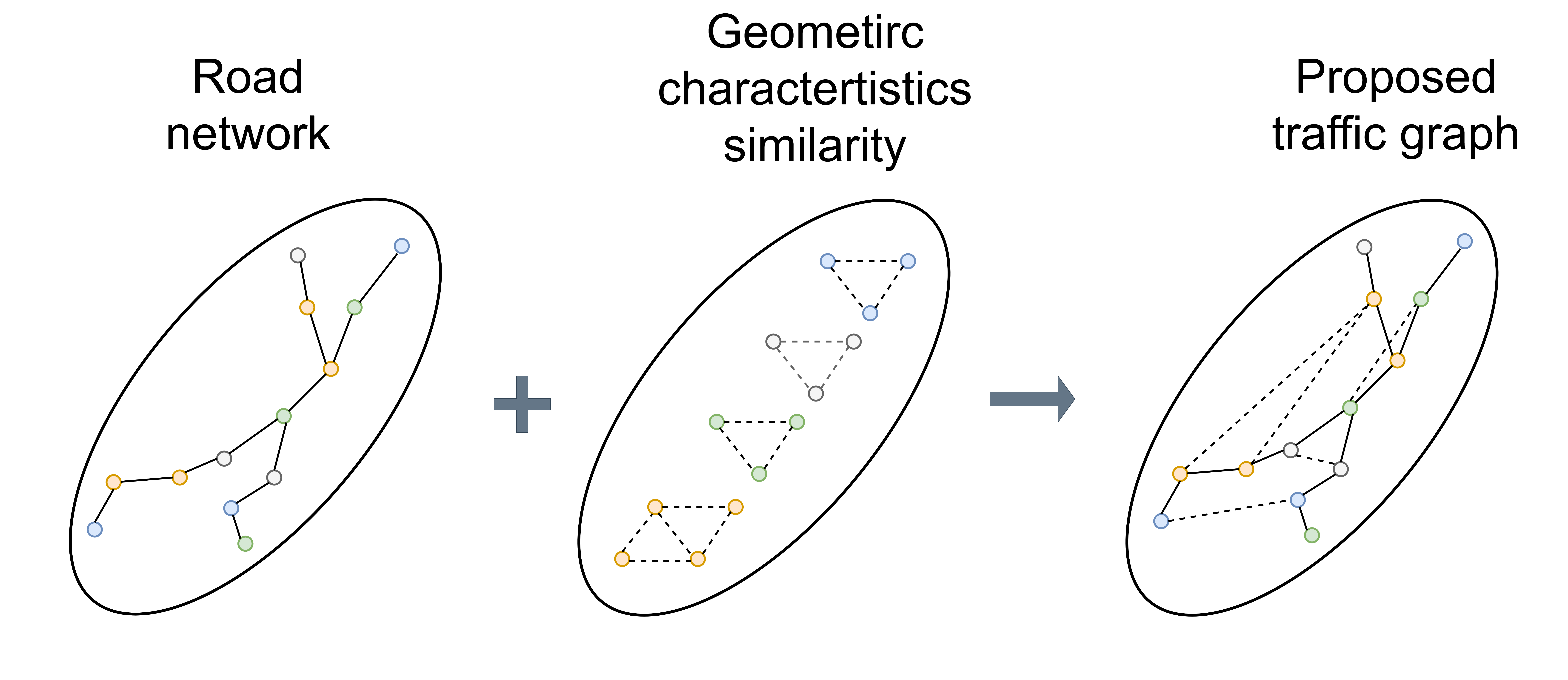}
  \caption{The graph design}
  \label{fig:topo}
\end{figure}

\subsection{Traffic graph construction}
In real world scenario, traffic modelling frequently encompasses considerations of locality and spatial correlations. Locality is often presented via geometric characteristics like adjacent roads, shape of the roadways, road class and types, which can notably impact the model performance \cite{Topo3}. Strong interactions among traffic nodes can occur due to the influence of those geometric characteristics even with distance and extend in bi-direction way \cite{ST-relation, Topo1, Topo2}. 
Hence, we consider those traffic POIs to better design a suitable graph.

In order to generate a suitable graph, we propose a strategy to fuse the geometric characteristics and road network, see \cref{fig:topo}. The initial graph is first built upon road network. Then, geometric characteristics are used to measure the similarity between nodes and form new edge connections if they are matched. Compared with the simple division based on the location or distance of traffic nodes, fusion approach can better retain variety of geometric-related influence by connecting similar nodes. The geometric characteristics we take into consideration are:

\begin{enumerate}
\item{Geographic location \cite{ST-relation, Fusion1}: the traffic measurements can be different regarding proximity to city centre and CBD, which generate distinct distribution and reduce the generalisability of modelling techniques.}
\item{Sub streets \cite{Topo4}: if the modelling region has sub streets, the traffic can be impacted by the incident differently.}
\item{Configuration of adjacent roads \cite{Topo3}: the configuration of modelling region can greatly impact the traffic features, for instance, an intersection merging additional lanes from adjacent roads will have higher volume and occupancy.}
\end{enumerate}

\subsection{Decentralised traffic incident detection with NL}
\label{subsec:decentralised-traffic-detection}

Suppose the traffic graph is defined, the NL-based optimisation framework as illustrated in \cref{fig:framework-1} will take this graph to initialise the undirected networked graph $G = (V, E)$ for optimisation purposes. 
Unlike centralised training, ADMM utilises a sound mathematical framework to decompose and solve the NL problem as sub-problems in parallel. Three key sub-problems are denoted as w-, z- and u-updates in \cref{alg:1}. In a data-decentralised scenario, w-update can be carried out within each distributed local nodes, while z- and u-updates can be run in the centralised server, see \cref{fig:framework-2}. Each node ${v}_{t} \in V$ can benefit from knowledge sharing which is baked into the optimisation routines executed as part of the aforementioned sub-problems. 
On the other hand, the private knowledge from local data was retained in $b$ in \cref{eq:1} during the optimisation process. Throughout the process, the detection model may produce accurate detection while allowing local data governance. This resonates well with real-world scenarios where vast number of traffic sensors are often sparsely deployed across different geographically-parted traffic regions.

Furthermore, while the proposed hierarchical framework appears as a predominantly client-server framework involving local nodes and a centralised server, it also promotes direct communication among neighboring nodes for knowledge sharing. This allows removing the minimal reliance on a centralised server, if needed, as demonstrated in \cite{NL1}.
In addition, the convex objective function of OC-SVM leads to the convex objective function for NL in \cref{eq:3}. Hence, ADMM is guaranteed to converge to a global optimum. Compared to FL where each node may have unaligned feature representations, NL-based framework promises guaranteed convergence, which bears the risk of divergence than FL.
\section{Evaluation}
\label{sec:experiment}

This section details out the experiments carried out in order to evaluate the performance of proposed framework, and report the results. In our experiments, we mainly focused on answering the following research questions:

\begin{itemize}
\item{RQ 1: How effective is the proposed approach for detecting traffic incidents in a \textit{context-aware manner} within a decentralised traffic network, which also includes incident rarity and incident-like patterns found in \textit{real-world} traffic scenarios.}
\item{RQ 2: How does the \textit{traffic graph design} impact the performance of proposed approach, and influence the training of OC-SVM models through knowledge sharing amongst \textit{similar traffic regions}?}
\item{RQ 3: How scalable the proposed approach is in the face of \textit{expanding traffic networks}?}
\end{itemize}

\subsection{Data}
\begin{figure}[h!]
  \centering
  \includegraphics[width=0.8\linewidth]{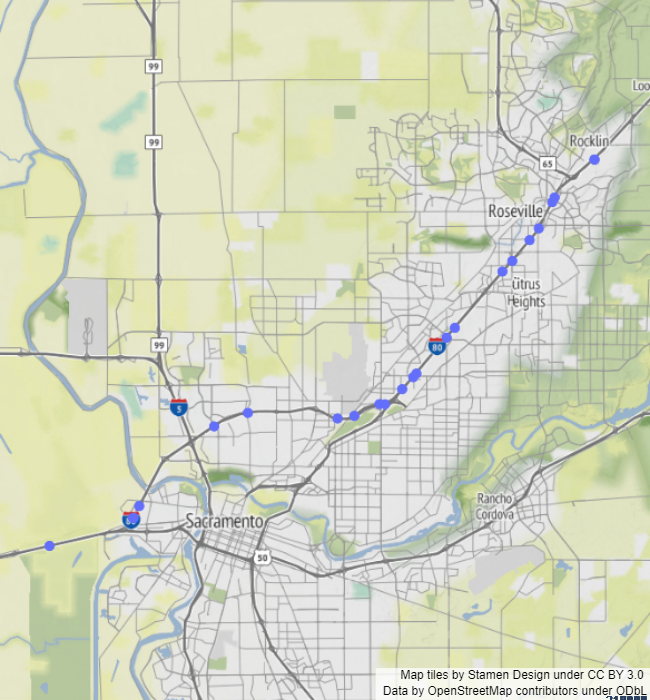}
  \caption{The selected region from California I-80 freeway, blue dots represent local nodes, created via open-street-map library}
  \label{fig:pems}
\end{figure}

The data used to train and test the proposed decentralised incident detection solution was sourced from the well-known California PeMS database \cite{DATA_PEMS}. This traffic incident dataset contains traffic data aggregated at 5-minute intervals and includes raw traffic features flow, speed, and occupancy collected from District 3 (Sacramento area) from January 1, 2016 to December 31, 2016. The incident data is also labelled with corresponding geometric location, start time and duration. To build a distributed traffic scenario, we deduced and selected 24 traffic regions from the I-80 freeway, which are under the influence of diverse geometric conditions (see Figure \cref{fig:pems}), and each was formed with 10 days (2640 records) incident-free samples for training. Due to the rarity of incidents, each local test set was curated with 5\% real incidents and 95\% unseen incident-free samples to simulate an imbalanced \textit{real-world} traffic scenario. This resulted in 1200 records (i.e. 60 incidents + 1140 unseen normal records) in total, distributed amongst all 24 traffic regions.

\subsection{Experimental settings}
The proposed framework was implemented using the python library CVXPy and trained in the OzStar supercomputer. The input size was fixed to 4 timesteps (i.e., 20 minutes) using the \textit{occupancy difference} between upstream and downstream sections to represent a practical traffic detection scenario \cite{review1}. Meanwhile, the following baseline models were also evaluated for comparison:
\begin{itemize}
    \item \textbf{Localised OC-SVMs:} a group of OC-SVMs trained within the \textit{local nodes} of individual traffic regions atop locally accumulated traffic data to represent a family of non-collaborative detection models.
    \item \textbf{Centralised OC-SVM:} a centralised OC-SVM trained with an aggregated training set consisting of all traffic incident data distributed across local nodes to represent a conventional centralised model.
    \item \textbf{Federated Autoencorder (FedAvg AE) \cite{Fed_AE}:} a representative FL approach that has recently been customised for incident detection, trained under similar experimental settings as that of the proposed.
\end{itemize}

For OC-SVM, the upper bound scaler $\nu$ was grid searched in range $(0.9, 1)$ to reflect the incident rarity, and used with a linear kernel function, based on observation. For NL, we ran the regularisation path of $\lambda$ described in \cite{NL1} with the search space of $(0, {10}^{+4})$, ${\epsilon}^{primal}$ and ${\epsilon}^{dual}$ are set as ${10}^{-3}$.  

\subsection{Evaluation metrics}

In all the experiments, we evaluated the models compared based on commonly used metrics in traffic incident detection tasks, which are Detection Rate (DR), False Alarm Rate (FAR), Area-Under-Curve (AUC) and F1-score. For the purpose of computing the F1-score, a True-Positive (TP) means the predicted incident does exist within the duration of the true incidents. A False-Positive (FP) is a predicted incident that is not a TP. A True-Negative (TN) means predicted sample is incident-free. Finally, a False-Negative (FN) is a true incident that was missed by the corresponding detection method. 
\begin{equation}
\label{eq:4}
    {t}_{i}^{detected} = 
    \begin{cases}
    {t}_{i}^{occured},& \text{if } {t}_{i} \text{ is detected}\\
    {t}_{i}^{duration},                & \text{otherwise}
    \end{cases}
\end{equation}

\begin{equation}
\label{eq:5}
    \textrm{adjusted MTTD} = \frac{1}{n} \sum_{i=1}^{n}({t}_{i}^{detected} - {t}_{i}^{occured})
\end{equation}

The reported results are filtered by 10\% FAR, using mean values over 10 runs. In addition, to address potential bias where a model with a low detection rate might have a quick detection of a few incidents, conflicting with metrics such as detection delay, we proposed the metric adjusted Mean-Time-To-Detect (ad-MTTD), calculated as shown in \cref{eq:5}, for a more accurate evaluation of detection delay. If the model miss the incident in detection, a penalty will be added to total detection time based on incident duration.

\subsection{Results}

\begin{table}[!ht]
    \centering
    \begin{tabular}{@{}lccccc@{}}
    \toprule
      Model&  ACC&  F1&  DR&  AUC& ad-MTTD \\
    \midrule
      Local &  0.633&  0.170&  0.750&  0.296& 2.950
\\
      Global &  0.913&  0.305&  0.383&  0.479& 7.483
\\
      Fedavg AE&  0.926&  0.546&  \textbf{0.890}&  \textbf{0.932}& \textbf{1.302}
\\
      (Our) NL &  \textbf{0.977}&  \textbf{0.791}&  \textbf{0.883}&  \textbf{0.946}& 2.083
\\
    \bottomrule
    \end{tabular}
    \caption{Results of the comparison between proposed and baseline methods with the method highlighted in bold if its performance is statistically the best one(s).}
    \label{tab:my_label_overall}
\end{table}


The results showed that the proposed approach which balance the global and local view leads to a more effective boundary that encapsulates the training samples from incidents' representation (see \cref{tab:my_label_overall}). The FAR for Localised OC-SVM fail to meet 10\% FAR in most attempts. Its low F1 and AUC score indicate localised OC-SVM is highly constrained by local samples, which is unreliable under a real-world standard. Centralised OC-SVM on the another hand, has a low detection rate when aggregating all training sets together. This could happen when traffic characteristics among nodes are unaligned with each other. The centralised model fitted with all training sets generates a large boundary to encapsulate diverse incident-free samples, which the learnt boundary could conceal the incidents that are detectable for individual nodes. 

\begin{figure}[h!]
  \centering
  \includegraphics[width=0.9\linewidth]{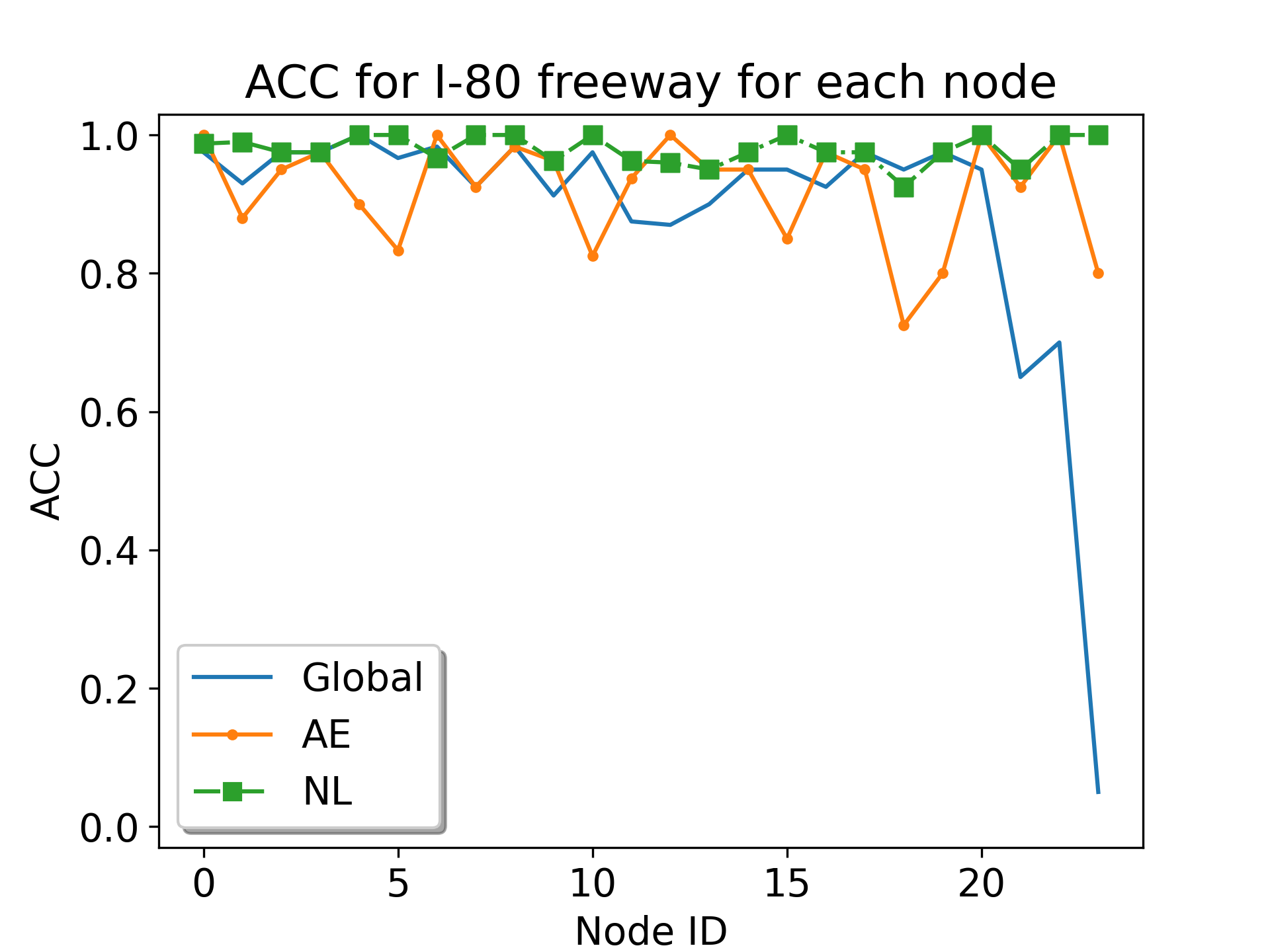}
  \caption{The accuracy metric for centralised OC-SVM, Fedavg AE and the proposed framework in each local node ordered by node id}
  \label{fig:acc_compare}
\end{figure}

\begin{figure}[h!]
  \centering
  \includegraphics[width=0.9\linewidth]{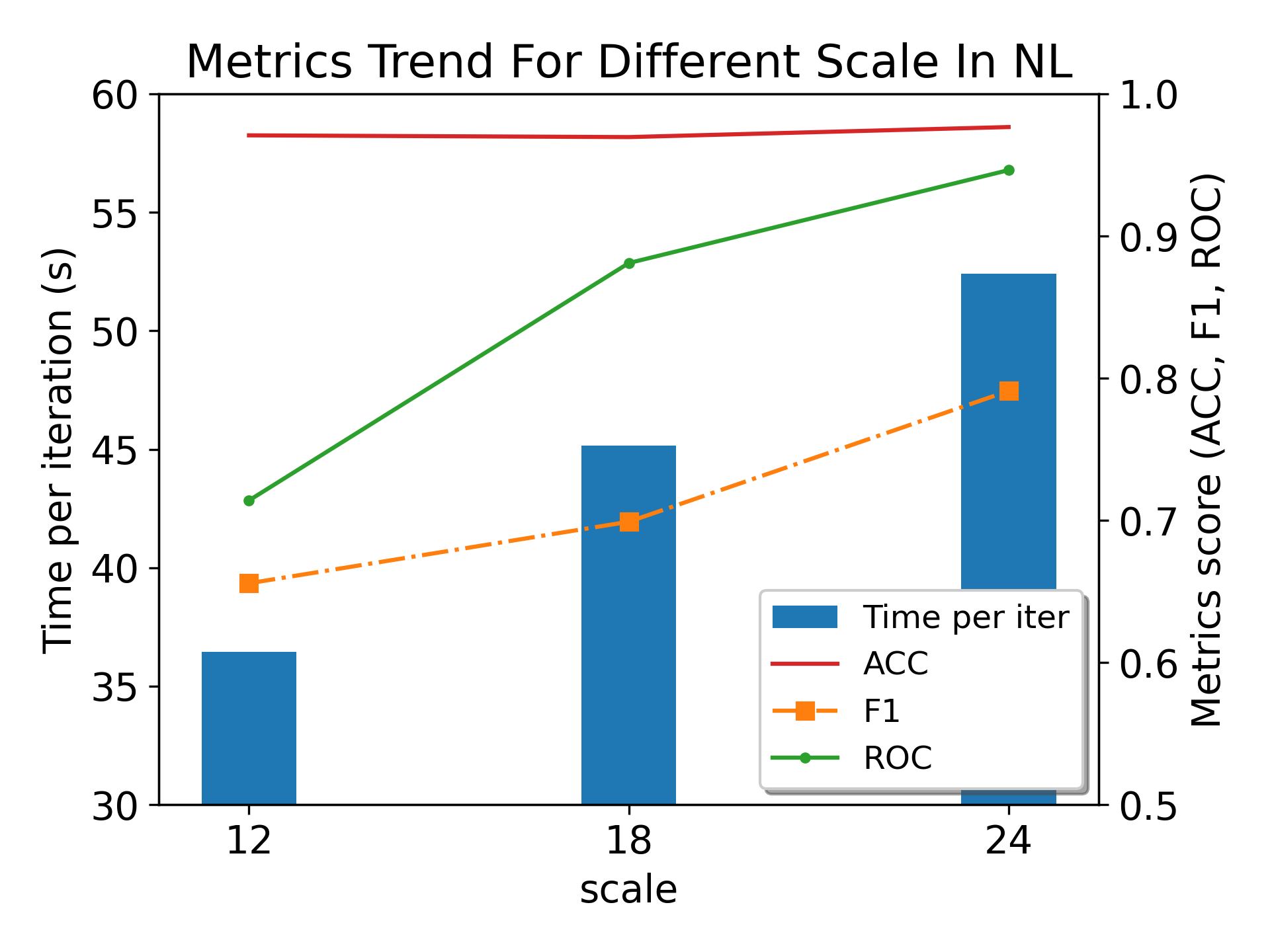}
  \caption{The performance metrics and computation time for different network scale using fused traffic graph in proposed framework}
  \label{fig:time_scale}
\end{figure}

In addition, our approach also outperformed the FedAvg AE on accuracy and F1-score, and showed competitive results in detection rate and AUC. Fedavg AE has the advantages of shorter ad-MTTD, which indicates a shorter delay for picking up incidents correctly. It was evident that our approach is able to perform competitively with respect to the more computationally-intensive autoencoder method \cite{Fed_AE2}. 
Meanwhile, our approach also exhibited consistent, robust performance across all individual nodes representing different traffic regions followed by FedAvg AE (see \cref{fig:acc_compare}). This success could be attributed to the ability of the NL-based proposed approach to \textit{leverage knowledge from other traffic regions} sharing similar \textit{context environments} for traffic incident detection fostering better generalised models. In comparison, Locally trained models tend to suffer from data scarcity, leading to under-performance issue. Furthermore, the centralised OC-SVM model also performed reasonably well on most nodes, but notably missed detecting incidents on node 24. Trained on an aggregated dataset with varying traffic characteristics, the centralised model can trigger false alarm on data instances which are normal for individual while out of the distribution from global view, just as the observed.


\subsection{Ablation study - impact of traffic graph design}

In our study, we employed a fusion strategy to construct the traffic graph for NL, to allow optimisation to happen in parallel based on the road network and geometric attributes of individual nodes.
To validate its effectiveness, we compared it with a pure road network-based and pure geometric characteristics-based graph designs. \cref{tab:ablation} reports the comparison results of these three strategies in terms of their eventually obtained testing metrics. 
Interestingly, the fusion strategy didn't notably enhance the accuracy, although it positively impacted all other metrics. 
This suggests that \textit{our fusion strategy effectively enhances the ability of localised models to distinguish misleading incident-like patterns from actual incidents, resulting in notable improvements in F1-score and detection rate}. Therefore, it can be deemed better suited to tackle the challenges exists in real-world traffic incident detection scenarios.

\begin{table}
    \centering
    \begin{tabular}{@{}lccccc@{}}
    \toprule
         Graph&  ACC&  F1&  DR&  AUC& ad-MTTD
\\
    \midrule
         Road&  0.968&  0.648&  0.583&  0.699& 4.067
\\
         Geo&  0.973&  0.736&  0.767&  0.851& 3.083
\\
         Fused&  \textbf{0.977}&  \textbf{0.791}& \textbf{ 0.883}&  \textbf{0.946}&\textbf{ 2.083}
\\
    \bottomrule
    \end{tabular}
    \caption{Results of the comparison between proposed, pure road network based and pure geometric characteristics based graph design with the design highlighted in bold if its performance is statistically the best one(s).}
    \label{tab:ablation}
\end{table}

\subsection{Scalability analysis}

We also evaluated the performance and time-to-converge of the proposed approach against a gradually increasing number of traffic regions (represented by local-nodes) within the underlying traffic network. Our results (see \cref{tab:my_label_scale}) showed that the detection rate and F1-score gained an incremental improvement as the number of traffic regions were gradually increased, which \textit{implies that larger the network scale is, the higher chance that localised models can benefit from information shared by neighboring nodes}. In addition, the metric \textit{time-to-converge also showed a favourable linear correlation with the network scale} (see \cref{fig:time_scale}). 
This relationship primarily stems from the increased number of traffic regions, leading to longer convergence times for the ADMM algorithm due to the tension introduced by collaborating nodes for knowledge sharing.

\subsection{Parameter sensitivity analysis}

\begin{figure}[h!]
  \centering
  \includegraphics[width=0.9\linewidth]{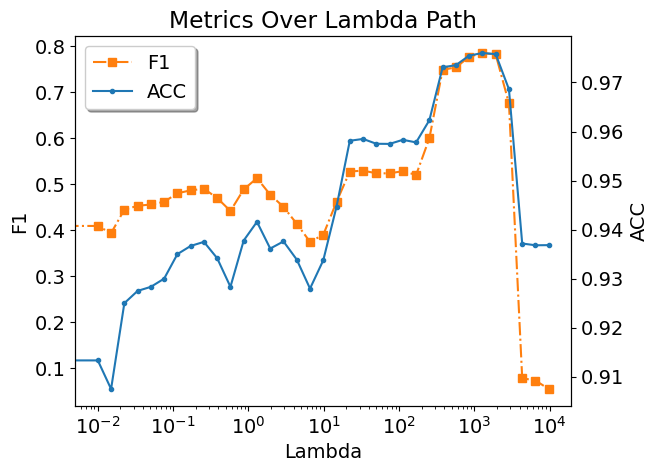}
  \caption{The regularisation path of $\lambda$ using fused traffic graph in proposed framework with network scale = 24}
  \label{fig:path}
\end{figure}

As discussed in \cref{sec:Preliminaries-NL}, the choice of $\lambda$ plays a pivotal role in the proposed NL-based approach. 
We conducted an analysis of $\lambda$ by gradually incrementing it, and evaluated the performance of the proposed scenario as in \cref{tab:my_label_overall}. The results, displayed in \cref{fig:path}, depict $\lambda$ in log-scale. At $\lambda = 0$ each node only uses its own training samples as it \textit{disables knowledge sharing amongst nodes}, 
and leads to a lower F1-score. When $\lambda$ was between $(200,400)$, the F1-score and overall accuracy steadily improved upon its optimal value, indicating that $\lambda_{critical}$ lies between $(200,400)$. It also represents the $\lambda$ where \textit{the algorithm has approximately split the nodes into their correct clusters representing similar context environments for traffic incident detection}, each with its own classifier \cite{NL1}.

\begin{table}[!ht]
    \centering
    \begin{tabular}{@{}lccccc@{}}
    \toprule
      Scale&  ACC&  F1&  DR&  AUC& ad-MTTD \\
    \midrule
      12&  0.965&  0.638&  0.611&  0.732& 4.806
\\
      18&  0.970&  0.699&  0.706&  0.881& 3.745
\\
      24&  \textbf{0.977}&  \textbf{0.791}&  \textbf{0.883}&  \textbf{0.946}& \textbf{2.083}
\\
    \bottomrule
    \end{tabular}
    \caption{Results of the comparison between proposed framework on different scale of the road network with the scale highlighted in bold if its performance is statistically the best one(s).}
    \label{tab:my_label_scale}
\end{table}

\section{Conclusions and future work}
\label{sec:conclusion}

In this study, we proposed an NL-based decentralised learning framework for traffic incident detection, capable of training conventional ML models on a local scale while leveraging other relevant data sources distributed elsewhere via the sharing and fusion of model parameters without data transfer. The relevance of local data sources is defined by a graph, where each node presents a local data source, and each edge reflects the relevance between its two connected local data sources. We designed a traffic data-specific graph through a fusion strategy, adopted OC-SVM as ML models, and used ADMM for model training. Experiments on a popular real-world traffic dataset validated the superiority of our approach in comparison to centralised and localised baselines. Compared to deep federated learning, our approach also demonstrated the competitive performance. In the future, we plan to study how to construct the data-relevance graph in an adaptive, data-driven but privacy-preserving manner, inspired by learning vector quantisation \cite{qin2005initialization} and graph matching \cite{gong2016discrete}, and investigate how to improve the scalability of the proposed method when facing a large amount of local data sources.

{
    \small
    \bibliography{main}
}

\end{document}